\documentclass[journal]{IEEEtran}
\usepackage{amsmath,amsfonts}
\usepackage{algorithmic}
\usepackage{algorithm}
\usepackage{array}
\usepackage[caption=false,font=normalsize,labelfont=sf,textfont=sf]{subfig}
\usepackage{textcomp}
\usepackage{stfloats}
\usepackage{url}
\usepackage{verbatim}
\usepackage{graphicx}
\usepackage{cite}
\usepackage{epsfig}

\usepackage{times}
\usepackage{amsthm}
\usepackage{amssymb}
\usepackage{amsopn}
\usepackage{amstext}
\usepackage{lipsum,multicol}
\usepackage{stfloats}
\usepackage{booktabs}
\usepackage{xfrac}
\usepackage{balance}
\usepackage{color}
\usepackage{mathtools}
\usepackage{tabularx}
\newcolumntype{L}{>{\raggedright\arraybackslash}X}
\newcolumntype{C}{>{\centering\arraybackslash}X}
\newcolumntype{R}{>{\raggedleft\arraybackslash}X}
\newcolumntype{P}[1]{>{\centering\arraybackslash}p{#1}}
\newcolumntype{H}[1]{>{\raggedright\arraybackslash}p{#1}}
\newcolumntype{T}[1]{>{\raggedleft\arraybackslash}p{#1}}

\usepackage{soul}
\usepackage{xcolor}
\sethlcolor{white}
\usepackage{empheq}
\usepackage{colortbl} 
\definecolor{lightyellow}{RGB}{255,255,200}

\newif\ifhighlight
\highlightfalse

\ifhighlight
  \sethlcolor{yellow}
\else
  \sethlcolor{white}
\fi

\soulregister\cite7
\soulregister\ref7

\usepackage{makecell}
\usepackage{pifont}
\newcommand{\circnum}[1]{\ding{\numexpr171+#1\relax}} 

\newcommand{\greenmark}{\color{green}\ding{52}}

\newcommand{\orangemark}{\color{orange}\ding{52}}

\newcommand{\redx}{\color{red}\ding{56}}

\usepackage{bm}
\usepackage{diagbox}
\usepackage{float}
\usepackage{epstopdf}
\usepackage{url}
\usepackage{multirow}
\usepackage[linkcolor=blue, citecolor=blue, urlcolor=blue,colorlinks=true]{hyperref}
\usepackage{threeparttable}
\usepackage{orcidlink}

\begin{document}

\title{\LARGE \bf
AC-LIO: Towards Asymptotic Compensation for Distortion in LiDAR-Inertial Odometry via Selective Intra-Frame Smoothing
}

\author{Tianxiang Zhang\textsuperscript{\orcidlink{0009-0001-4784-4548}}, Graduate Stutent Member, IEEE, Xuanxuan Zhang\textsuperscript{\orcidlink{0009-0007-1209-5712}}, Graduate Stutent Member, IEEE, \\ Wenlei Fan\textsuperscript{\orcidlink{0009-0005-3481-6918}}, Xin Xia\textsuperscript{\orcidlink{0000-0002-5108-7578}}, Senior Member, IEEE, Huai Yu\textsuperscript{\orcidlink{0000-0001-6043-3412}}, Member, IEEE, Lin Wang\textsuperscript{\orcidlink{0000-0002-7485-4493}}, Member, IEEE,\\ and You Li\textsuperscript{\orcidlink{0000-0003-3785-0976}}, Senior Member, IEEE %
\vspace{-1em}
\thanks{
This work is partly supported by the National Natural Science Foundation of China (42274052) and the Science and Technology Plan of Shenzhen under Grant (JCYJ20240813151253068).
}%
\thanks{$^{1}$T. Zhang, X. Zhang, W. Fan, and Y. Li are with State Key Laboratory of Information Engineering in Surveying, Mapping and Remote Sensing (LIESMARS), Wuhan University, China. 
X. Zhang is the corresponding author. Email: \tt \footnotesize \{\small cyberkona; xuanxuanzhang; wenleifan; liyou\footnotesize \}\small @whu.edu.cn }%
\thanks{$^{2}$Xin Xia is with Department of Mechanical Engineering, University of Michigan-Dearborn. Email: \tt \small xinxia@umich.edu}%
\thanks{$^{3}$H. Yu is with Electronic Information School, Wuhan University, China. Email: \tt \small yuhuai@whu.edu.cn}%
\thanks{$^{4}$T. Zhang, L. Wang are with School of Electricial and Electronic Engineering, Nanyang Technological University, Singapore. Email: \tt \small n2409916c@e.ntu.edu.sg; linwang@ntu.edu.sg}%
}


\maketitle

\begin{abstract}
Existing LiDAR-Inertial Odometry (LIO) methods typically utilize the prior trajectory derived from {IMU integration} to compensate for the motion distortion within LiDAR frames.
However, discrepancies between {the prior and the true trajectory} can lead to residual motion distortions that compromise the consistency of {LiDAR frames with their corresponding} geometric environment.
This imbalance may result in pointcloud registration becoming trapped in local optima, thereby exacerbating drift during long-term and large-scale localization.
To this end, {this work proposes} a novel LIO framework with selective intra-frame smoothing dubbed AC-LIO.
The core idea is to asymptotically backpropagate {the} current update term and compensate for residual motion distortion under the guidance of {a} convergence criterion, aiming to improve the accuracy of {discrete-state LIO systems} with minimal computational increase.
Extensive experiments demonstrate that {the} AC-LIO framework further enhances odometry accuracy compared to {state-of-the-art methods}, with {an approximately 28.5\% reduction} in average RMSE over the {second-best} result, leading to marked improvements in the accuracy of long-term and large-scale localization and mapping.
{The} code and demo are available {at} \url{https://cyberkona.github.io/publication/ac-lio/}.
\end{abstract}


\section{Introduction}
\IEEEPARstart{W}{ith} {advances} in Simultaneous Localization and Mapping (SLAM) technology, robots are increasingly achieving precise perception of external environments and their own motion\cite{taheri2021slam, cadena2016past, chen2020active}. 
As a vital subset of SLAM, LiDAR-Inertial Odometry (LIO) has gained widespread application in autonomous driving~\cite{wang2021intensity}, unmanned logistics~\cite{duan2022stereo} and other fields, {due to} its high robustness and accuracy~\cite{wang2024givl, lv2023continuous}.
Compared to Visual-Inertial Odometry, {the} LIO system can directly obtain depth observations of {the} surroundings, without the need to estimate the depth of feature points~\cite{jin2020camera}, and exhibits better adaptability to different lighting conditions~\cite{han2023camera, song2023ir}.

\begin{figure}[t]
    \begin{center}
        {\includegraphics[width=1\columnwidth]{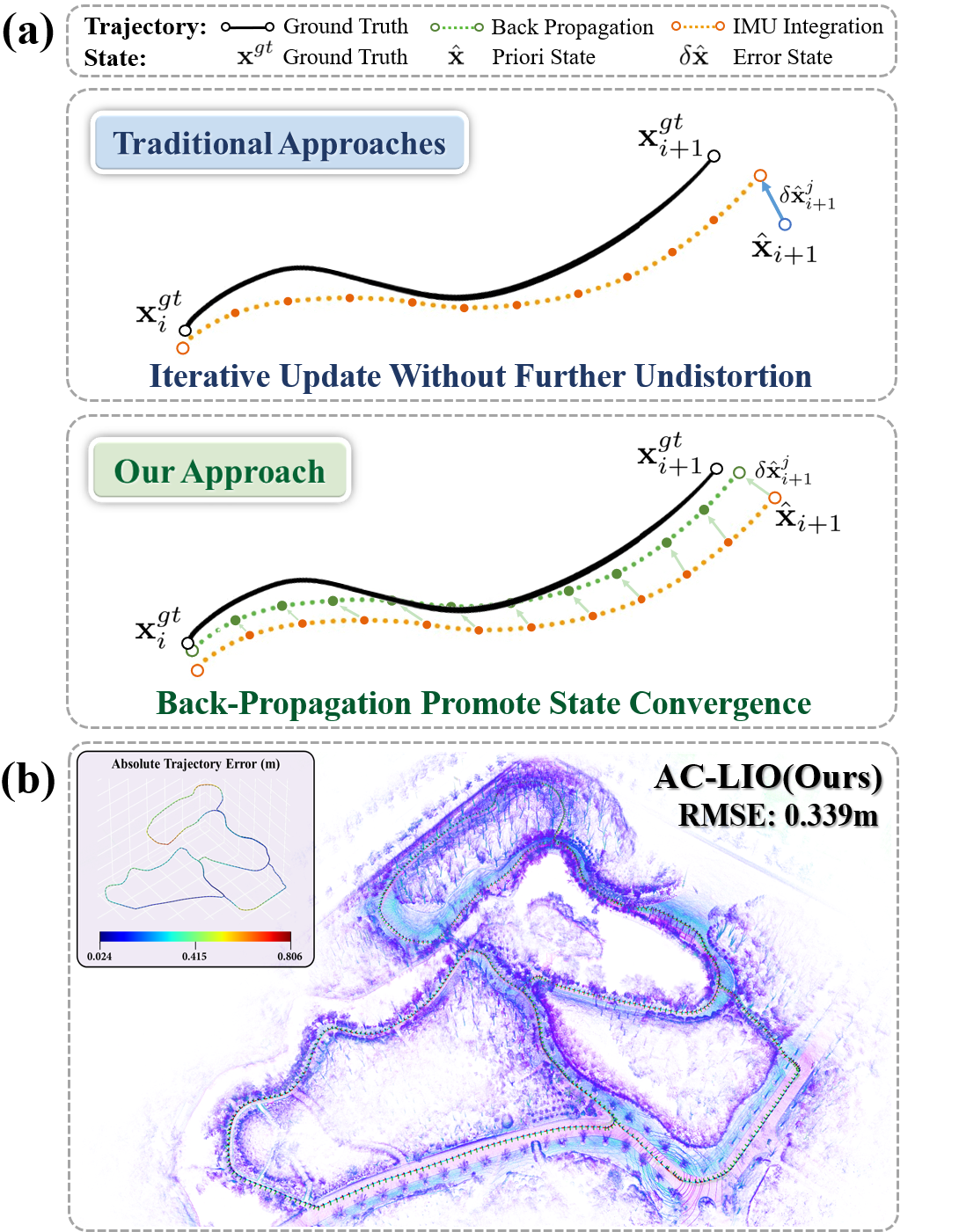}}
    \end{center}
    \vspace{-0.3cm}
    \caption{\label{fig:intro_part}\textbf{(a)} \textbf{A {conceptual} comparison of AC-LIO with traditional approaches.} Conventional LIO typically considers an initial distortion compensation via IMU, yet the residual motion distortion may prevent the registration from further convergence. AC-LIO employs a selective intra-frame smoothing strategy to asymptotically backpropagate and compensate for residual distortion. \textbf{(b)} \textbf{Large scale localization and mapping result of BotanicGarden dataset.} {AC-LIO} achieves {better accuracy} than other benchmark frameworks on the sequence Garden\_1008 with VLP16 LiDAR.}
    \vspace{-0.4cm}
\end{figure}

\begin{table*}[b]
\vspace{-2mm}
\caption{Comparison of Representative LO/LIO Frameworks {for Motion} Undistortion}
\vspace{-1mm}
\renewcommand{\arraystretch}{1.25}
\centering
\resizebox{1\linewidth}{!}{
\label{tab:motion_distortion_methods}
\begin{tabular}{cccccc}
\toprule[1pt]
\textbf{Category} & \textbf{Method} & \textbf{Year} & \textbf{Initial Undistortion} & \textbf{Further Undistortion} & \textbf{Extra States/Estimators} \\
\midrule

\multirow{5}{*}{\rotatebox{90}{\scriptsize \shortstack{ \textbf{(A) LiDAR-Only} \\ \textbf{Odometry}}}} 
 & LOAM~\cite{zhang2014loam} & 2014 & Constant Velocity & {\redx} & - \\
 & F-LOAM~\cite{wang2021f} & 2021 & Constant Velocity & {\redx} (After Convergence) & - \\
 & CT-ICP~\cite{dellenbach2022ct} & 2022 & - & {\greenmark} Two-State Interpolation & {\greenmark} Start and End States \\
 & I$^2$EKF-LO~\cite{yu2024i2ekf} & 2024 & Constant Velocity & {\greenmark} Process Noise Estimation & {\orangemark} Extra Iteration for Prediction  \\
 & Traj-LO~\cite{zheng2024trajlo} & 2024 & Smoothness Constraint & {\greenmark} Slide Window Optimization & {\orangemark} Multiple Control States \\

\midrule
\multirow{7}{*}{\rotatebox{90}{\scriptsize \shortstack{\textbf{(B) Discrete} \\ \textbf{State LIO}}}}  
 & LIO-SAM~\cite{shan2020lio} & 2020 & IMU Preintegration & {\redx} & (Preintegrated Factor) \\
 & FAST-LIO2~\cite{xu2022fast} & 2022 & IMU Integration & {\redx} & - \\
 & S.E.-LIO~\cite{yuan2023semi} & 2023 & IMU Preintegration & {\redx} & (Two-State Factor) \\
 & Point-LIO~\cite{he2023point} & 2023 & {(Not Necessary)} & {\greenmark} Point-by-point Update & {\orangemark} Per-point States \\
 & SR-LIO~\cite{yuan2024sr} & 2024 & Segment Integration & {\redx} & (Reconstructed Frame States) \\
 & AS-LIO~\cite{zhang2024lio} & 2024 & Bi-directional Integration & {\redx} & - \\
 & V.A.-LIO~\cite{dong2025vibration} & 2025 & IMU Integration & {\redx} & (Post-undistortion Uncertainty) \\

\midrule
\multirow{4}{*}{\rotatebox{90}{\scriptsize \shortstack{\textbf{(C) Continuous} \\ \textbf{State LIO}}}}  
 & CLINS~\cite{lv2021clins} & 2021 & B-spline Initialization & {\greenmark} Continuous-time B-spline & {\orangemark} B-spline Control Points \\
 & SLICT~\cite{nguyen2023slict} & 2023 & IMU Propagation & {\greenmark} {Propagation on Estimate} & {\orangemark} {Intermediate States} \\
 & DLIO~\cite{chen2023direct} & 2023 & Coarse-to-fine Deskew & {\redx} & (Continuous Analytical Model) \\
 & Traj-LIO~\cite{zheng2024trajlio} & 2024 & Gaussian Process Interpolation & {\greenmark} Slide Window Optimization & {\orangemark} Multiple Interpolated States \\

\midrule
\textbf{Ours} & \textbf{AC-LIO} &  & \makecell{IMU Integration \\ (with Information Storage)} & \makecell{{\greenmark} Selective Intra-frame \\ Asymptotic Compensation} & {\greenmark} (No Extra States/Estimators) \\

\bottomrule[1pt]
\end{tabular}
}
\vspace{-1em}
\end{table*}
\vspace{1mm}
As {the} LIO system continues to evolve, there still remain several fundamental issues to be addressed~\cite{lee2024lidar, li2021towards, liao2025real}. One of them is the conflict between continuous motion and rigid pointcloud registration~\cite{mcdermott2023correcting}.
Since the acquisition of LiDAR frames is often accompanied with the ego-motion of the vehicle, the LiDAR points with different timestamps actually lie in different coordinate systems.

As shown in Fig.~\ref{fig:intro_part}, traditional methods~\cite{xu2021fast, xu2025dynamic, wu2022robust} typically only consider using IMU integration for the initial compensation of motion distortion, without further optimization of the residual distortion in the state estimation.
This residual motion distortion will compromise the consistency between the LiDAR frame {and} its corresponding geometric environment, and it grows with increasing motion non-linearity.
Therefore, in conventional LIO systems, such residual error may suppress further convergence of pointcloud registration during the iterative update, ultimately impacting the odometry accuracy and causing severe drift.

However, few studies {have leveraged} the Rauch-Tung-Striebel (RTS) smoother to further optimize motion distortion within discrete-state LIO frameworks.
While smoothing strategies have been widely adopted in continuous state estimation and several sliding window factor graph frameworks, they primarily focus on joint optimization across multiple states, which often leads to substantial computational consumption.

{To balance real-time performance and accuracy in LIO systems, this paper proposes AC-LIO, a novel discrete-state LIO framework with selective intra-frame smoothing.
Without introducing additional states or estimates, AC-LIO leverages} smoothing methods to achieve asymptotic compensation for residual motion distortion during iterative state estimation.
Moreover, some redundant {computations can be avoided} as the state covariance chains are available from IMU prior propagation.
{AC-LIO} introduces two key technical contributions. 

First, during iterative state estimation, the proposed \textit{asymptotic distortion compensation}~(Sec.~\ref{asymptotic compensation}) method can backpropagate the current update term based on the prior state chain, and asymptotically compensate for the residual distortion.
This approach could enhance the consistency between the current LiDAR frame and the represented geometric environment, and avoid the pointcloud registration from falling into local {optima} due to the residual distortion, thereby promoting the accuracy of long-term and large-scene localization.

Second, due to the lack of correlation between initial error and motion distortion of current frame, {the \textit{convergence criterion}~(Sec.~\ref{convergence criterion}) is proposed based on the point-to-plane residuals to control the back propagation.}
The criterion is applied to assess the convergence of the pointcloud registration and ensure that the backpropagation and asymptotic compensation only happens under the condition of sufficient convergence in previous LiDAR frame, hence guaranteeing the robustness of {the} LIO system.
Compared to other strategies such as continuous-time estimation or frame segmentation, {AC-LIO leverages} this criterion to construct a feedback mechanism that promotes state convergence without additional estimation, achieving a balance between localization accuracy and computational efficiency~(Sec.~\ref{time efficiency}). It can be seamlessly embedded into the iterative Kalman Filter framework, offering great flexibility and scalability.

In summary, the main contributions are as follows: 

\begin{itemize}

\item A \textit{novel} framework AC-LIO with \textit{asymptotic distortion compensation} is proposed to further suppress residual distortion during iterative update, promoting system accuracy in long-term and large-scale scenarios.

\item The \textit{convergence criterion} based on point-to-plane residuals is introduced to evaluate the convergence and guide backpropagation.
Backpropagation is only performed upon sufficient convergence of the previous registration, thus further enhancing the system robustness.

\item Compared to other optimization strategies for motion distortion, the AC-LIO framework introduces \textit{no extra estimates or states}, making the asymptotic compensation less computationally expensive but highly flexible.

\item  Extensive experiments on AC-LIO are conducted across a series of real-world datasets with various scenarios. 
The results demonstrate that the AC-LIO framework achieves higher accuracy compared to current state-of-the-art frameworks. The code will be released to support and contribute to the open-source community.

\end{itemize}
\vspace{0.5mm}
\section{RELATED WORK}

\subsection{LiDAR-Only Odometry}
Studies such as LOAM~\cite{zhang2014loam} and F-LOAM~\cite{wang2021f} {have achieved} significant performance with the assumption of constant velocity motion during the corresponding {LiDAR frame intervals}.
However, this assumption {holds} effectively when {the frame} interval is {short enough and the motion} is slow and smooth. 
In case of drastic velocity {variations}, {the constant velocity model fails to} accurately characterize current motion state, leading to increased drift and even odometry divergence.
CT-ICP~\cite{dellenbach2022ct} {optimizes} the start and end states of LiDAR frame, and {constructs a continuous-time trajectory through interpolation}. {However,} the real movement may not follow the interpolated trajectory, {and it is difficult to determine the logical constraints between adjacent frames}.
I$^2$EKF-LO~\cite{yu2024i2ekf} {uses additional prediction iterations} to estimate motion distortion, {but} the update of process noise still depends on some hyperparameters.
Traj-LO~\cite{zheng2024trajlo} {exploits the} high frequency of LiDAR points and employs {multiple linear segments within each frame, enabling} more accurate trajectory characterization and suppression of motion aberrations during {high-speed} motion, but also introduces {a higher} computational cost.

\subsection{Discrete State LiDAR-Inertial Odometry} 
For LiDAR-inertial odometry, the IMU observation {provides} relatively reliable prior state information. 
FAST-LIO2~\cite{xu2022fast} {uses} the IMU state backpropagation to {undistort the pointcloud, followed by} state estimation through Error State Kalman Filter.
LIO-SAM~\cite{shan2020lio} {employs} the IMU pre-integration to de-skew pointcloud, {fusing} multiple constraints {from} LiDAR, IMU, and GPS through {a} factor graph.
{In subsequent LIO frameworks}, SR-LIO~\cite{yuan2024sr} {introduces} a LiDAR frame reconstruction strategy, while AS-LIO~\cite{zhang2024lio} {utilizes} an adaptive sliding window guided by spatial overlap degree. {Both approaches enhance the update frequency and reduce the impact of motion distortion, but lack an explicit mechanism to actively compensate for the residual distortion.}
Semi-Elastic-LIO~\cite{yuan2023semi} replaces the interpolation model with IMU motion constraints on the basis of CT-ICP framework, {improving} the stability of the inter-frame logical constraints.
Vibration-aware LIO~\cite{dong2025vibration} introduces post-undistortion uncertainty {to mitigate inaccuracies and improve reliability} under intense vibration conditions.
Although these methods {mitigate the effect} of pointcloud distortion to some extent, {further optimization of residual motion distortion remains an unresolved challenge}.
Point-LIO~\cite{he2023point} adopts a point-by-point update strategy to avoid the motion distortion issue and {achieve} extremely high-frequency odometry. Compared to {frame-based} registration, {this strategy is more susceptible to dynamic objects and may degenerate in adverse scenarios}.

\subsection{Continuous State LiDAR-Inertial Odometry} 
To characterize trajectories more precisely in state estimation, {several frameworks have proposed} new optimization solutions.
SLICT~\cite{nguyen2023slict} employs continuous-time factors to {model} LiDAR constraints within a sliding window. CLINS~\cite{lv2021clins} and SLICT2~\cite{nguyen2024eigen} introduce the B-spline for continuous state estimation.  
However, {unconstrained trajectories may not} always conform to {the} predefined model due to {limited time resolution}, {especially in high-speed dynamic motion}. Moreover, the continuous-time optimization strategy {can introduce higher computational costs}.
DLIO~\cite{chen2023direct} uses analytical equations to model {continuous trajectories} for precise motion correction after coarse deskewing with IMU integration. Yet it remains limited by {the inherent accuracy of the IMU} and neglects further undistortion {in the back-end registration}.
Traj-LIO~\cite{zheng2024trajlio}, developed from Traj-LO, leverages Gaussian Process to predict {continuous-time trajectories and maintain multiple states} through interpolation. {This avoids over-reliance on the IMU and enables sliding window optimization}.

In contrast, the proposed AC-LIO framework utilizes {a} convergence criterion to guide the backpropagation and asymptotic compensation for residual motion distortion, {thereby} improving the accuracy {in} long-term and large-scale scenarios while {maintaining} the system robustness. Table~\ref{tab:motion_distortion_methods} provides a comparison of representative LO/LIO frameworks for motion undistortion.

\section{METHODOLOGY}

\subsection{System Overview}

\begin{figure*}[t!]
    \vspace{2mm}
    \begin{center}
        {\includegraphics[width=2\columnwidth]{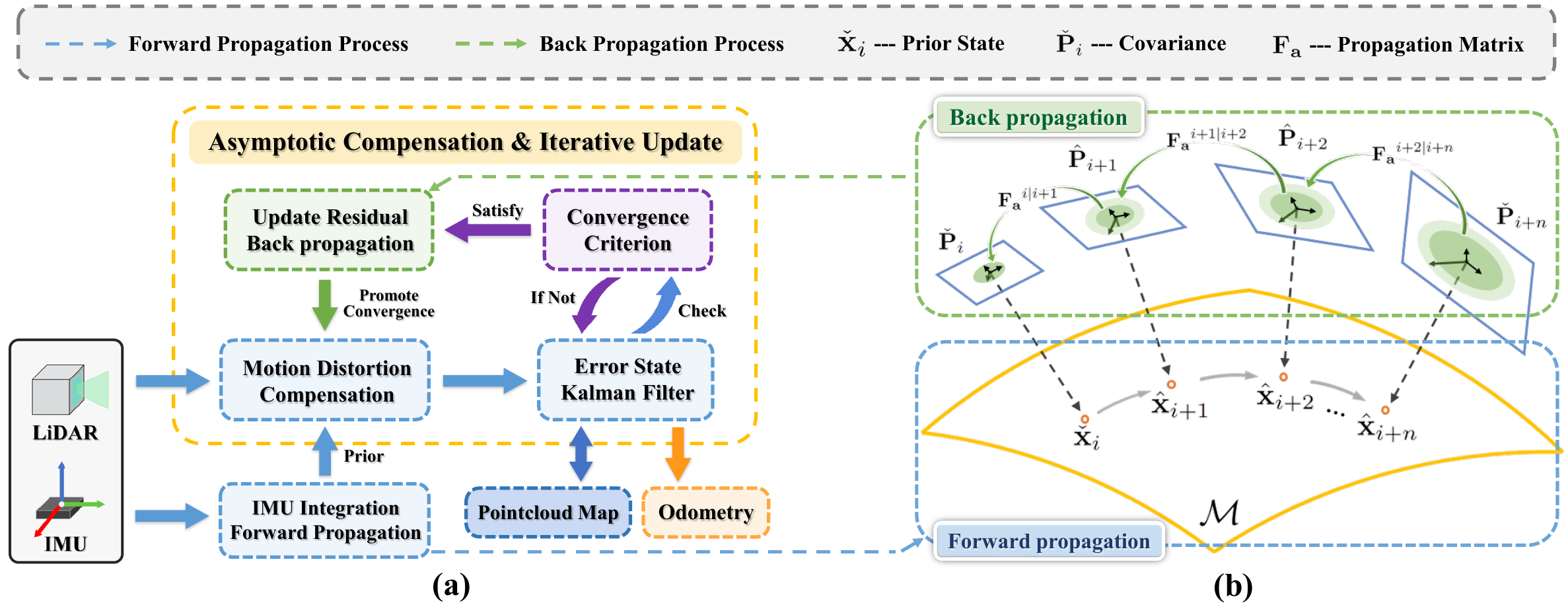}}
    \end{center}
    \vspace{-5mm}
    \caption{\label{fig:System Overview}\textbf{(a)} \textbf{The system overview of the AC-LIO framework}. During iterative ESKF, the back propagation and asymptotic compensation for pointcloud distortion are conducted based on the convergence criterion. \textbf{(b)} \textbf{On-manifold propagation}. The blue {denotes} the forward propagation of IMU integration, {while the} green indicates the backward recursion of the update term according to the propagation matrix and noise relation of error state chains.}
    \vspace{-4mm}
\end{figure*}

The system overview of {the} AC-LIO framework is shown in Fig.~\ref{fig:System Overview}, and the important notations {are} shown in Table~\ref{tab:notations}.
First, the IMU observations are fed into the forward propagation module to obtain prior trajectories.
{Then, the} LiDAR data utilize the prior state chain for initial compensation and {are sent} to the Error State Kalman Filter (ESKF) for state estimation~\cite{sola2012quaternion}, where the average point-to-plane residual (APR) is calculated after each update.

During the iterative ESKF, if the APR of previous LiDAR frame is lower than the convergence criterion and the current APR exceeds it, then the backpropagation of update residual will be triggered.
After each backpropagation, the LiDAR frame will be re-compensated based on the updated states to promote the convergence of registration. 
Otherwise, if the conditions are not met, the system will skip the backpropagation and continue the remaining iterations.
Once reaching the {maximum number} of iterations, the system will merge the LiDAR frame under the posterior pose into pointcloud map and publish the current odometry.

\subsection{ESKF State Estimation}

\begin{table}[t]
\renewcommand{\arraystretch}{1.05}
\footnotesize
\centering
\vspace{-1mm}
\caption{Important Notations}
\vspace{-1mm}
\label{tab:notations}
\begin{tabularx}{0.95\linewidth}{@{}lX@{}}
\toprule
\textbf{Symbols} & \textbf{Definition} \\
\midrule

$\boxplus/ \boxminus$ 
    & The boxplus and boxminus operators on manifold. \\
$I, L, G$
    & The IMU, LiDAR, and global coordinate frame. \\
$\mathbf{x}^{gt}$
    & The ground truth of state $\mathbf{x}$. \\
$\mathbf{x}_i$
    & The vector of state $\mathbf{x}$ at timestamp $i$. \\
$\hat{\mathbf{x}}$
    & The priori estimation of state $\mathbf{x}$. \\
$\hat{\mathbf{x}}^j, \delta \hat{\mathbf{x}}^j$
    & The $j$-th update of state $\mathbf{x}$ and error state $\delta\mathbf{x}$. \\
$\bar{\mathbf{x}}, \check{\mathbf{x}}$
    & The updated state $\mathbf{x}$ with or without back propagation. \\
$\mathbf{F_p}^{i|i+n}$
    & The propagation matrix from state $i$ to $i+n$.\\
$\mathbf{G}_{i}^{i+n}$
    & The backward gain matrix of state $i$ with respect to $i+n$.\\
$\mathbf{z}, z$
    & The vector and its scalar value of point-to-plane residual. \\
$\phi$
    & The incident angle of single laser beam on the fitted plane. \\  
$\sigma_t$
    & The expectation of average point-to-plane residual. \\
\bottomrule

\end{tabularx}
\vspace{-3mm}
\end{table}

For efficiency and accuracy concerns, the proposed AC-LIO utilizes the ESKF framework for state estimation. Based on the kinematic model and IMU measurement model, the state $\mathbf{x}$ , input $\mathbf{i}$, and noise $\mathbf{n}$ can be defined as follows:
\begin{equation}
    \label{equation1}
    \begin{split}
        \mathbf x & \doteq \begin{bmatrix} \mathbf{R}_I^G & \mathbf{p}_I^G & \mathbf{v}_I&\mathbf{b}_{\bm \omega} & \mathbf{b}_{\mathbf a} & \mathbf{g}^G \end{bmatrix}^T \\
        \mathbf i & \doteq \begin{bmatrix} {\bm \omega}_m & {\mathbf a}_m\end{bmatrix}^T,~~
        \mathbf n \doteq \begin{bmatrix}\mathbf n_{\bm \omega}&\mathbf n_{\mathbf a}&\mathbf n_{\mathbf b\bm \omega}&\mathbf n_{\mathbf b\mathbf a}\end{bmatrix}^T 
    \end{split}
\end{equation}

Where $\mathbf{R}_I^G$, $\mathbf{p}_I^G$ and $\mathbf{v}_I^G$ {denote the IMU orientation}, position and linear velocity in $\mathit{G}$, and $\mathbf{g}^G$ is the local gravity. 
The $\mathbf{n}_{\bm \omega}$, $\mathbf{n}_{\mathbf a}$ are {noise terms} of angular velocity ${\bm \omega}_m$ and linear acceleration ${\mathbf a}_m$, the $\mathbf{b}_{\bm \omega}$, $\mathbf{b}_{\mathbf a}$ are {IMU biases} with $\mathbf {n}_{\mathbf{b}\bm \omega}$ and $\mathbf n_{\mathbf b\mathbf a}$ as their noises in Gaussian-Markov model.
Since $\mathbf{x}$ can be regarded as a high-dimensional manifold locally homeomorphic to Euclidean space $\mathbb{R}${, the} operations $\boxplus$ and $\boxminus$ on manifold can be defined according to~\cite{hertzberg2013integrating}. 
Define the error state as $\delta\mathbf{x} \doteq \mathbf{x}^{gt} \boxminus \hat{\mathbf{x}}$, and the recursions for $\delta\mathbf{x}$ and the covariance $\mathbf{P}$ can be expressed as follows:
\begin{equation}
    \label{equation2}
    \begin{gathered}
        \quad \quad \delta \mathbf{x}_i  = \begin{bmatrix} \delta {\bm \theta} & \delta \mathbf{p}_I^G & \delta \mathbf{v}_I& \delta \mathbf{b}_{\bm \omega} & \delta \mathbf{b}_{\mathbf a} & \delta \mathbf{g}^G \end{bmatrix}^T \\
        \quad \delta \hat{\mathbf{x}}_{i+1} = \mathbf{F_p} \delta \check{\mathbf{x}}_{i} + \mathbf{F_n} \mathbf{n}_i \\
        \quad \hat{\mathbf{P}}_{i+1} = \mathbf{F_p} \check{\mathbf{P}}_{i} \mathbf{F_p}^T + \mathbf{F_n} \mathbf{Q} \mathbf{F_n}^T
    \end{gathered}
\end{equation}

Assume the extrinsic $\mathbf{T}_{L}^{I}$ between LiDAR and IMU is already obtained from calibration.
For a point $\mathbf{p}$ in LiDAR frame, its neighboring cluster can be searched by KNN {algorithm} in pointcloud map. If a plane $S$ can be fitted to this cluster, then the normal vector $\mathbf{u}$ of this plane can be obtained. 
Note one point of the cluster as $\mathbf{q}$ and $( \cdot )_{\wedge}$ is the upper triangular matrix of vector, then the point-to-plane residual $\mathbf{z}$ and measurement matrix $\mathbf{H}$ are as follows:
\begin{equation}
    \label{equation3}
    \begin{gathered}
        \mathbf{z} = \mathbf{u}(\hat{\mathbf{p}}^G - \mathbf{q}^G)^T \\
        \mathbf{H} = \begin{bmatrix} -\mathbf{u}\hat{\mathbf{R}}_I^G(\mathbf{T}_{L}^{I} \mathbf{p}^L)_{\wedge} & \mathbf{u} & \mathbf{0}_{3 \times 12} \end{bmatrix}
    \end{gathered}
\end{equation}

Considering the nonlinear errors in the $\mathbf{F_p}$, $\mathbf{F_n}$ and $\mathbf{H}$, the estimation may require iteration to promote convergence. 
Note that $\mathbf{A}$ is an intermediate matrix in $SE(3)$ logarithmic map, as defined in~\cite{murray1995proportional}, and $\mathbf{B}^j$ denotes the Jacobian of update component with respect to $\delta \mathbf{x}$ during $j$th iteration, given by:
\begin{equation}
    \label{equation4}
    \begin{gathered}
        \mathbf{A} (\mathbf{u})^{-1} = \mathbf{I} - \frac{1}{2} (\mathbf{u})_{\wedge} + \left(1-\bm{\alpha} (\Vert \mathbf{u} \Vert\right)) \frac{\left(\mathbf{u}\right)_{\wedge}^2}{\Vert\mathbf{u}\Vert^2} \\
        \bm{\alpha} (\rm{\psi}) = \frac{\rm{\psi}}{2} {\rm cot}(\frac{\rm{\psi}}{2}) \\
        \mathbf{B}^j \!=\!\begin{bmatrix} \mathbf A\left(\delta {\bm \theta}^j \right)^{-T} \!\!& \mathbf 0_{3\times 15} \\
        \mathbf 0_{15\times 3} & \mathbf I_{15\times 15} \end{bmatrix}
    \end{gathered}
\end{equation}

According to the definition in~\cite{xu2021fast}, substituting the linearization of the prior and the measurement distribution into the maximum a-posteriori (MAP) estimate yields the following update equation:
\begin{equation}
    \label{equation5}
    \begin{gathered}
        \mathbf{K} = \mathbf{P}\mathbf{H}^T(\mathbf{H}\mathbf{P}\mathbf{H}^T + \mathbf{R})^{-1} \\
        \hat{\mathbf{x}}_{i+1}^{j+1} = \hat{\mathbf{x}}_{i+1}^{j} \boxplus [ -\mathbf{K}\mathbf{z} - (\mathbf{I} - \mathbf{K}\mathbf{H}) (\mathbf{B}^j)^{-1} (\hat{\mathbf{x}}_{i+1}^{j} \boxminus \hat{\mathbf{x}}_{i+1} ) ]
    \end{gathered}
\end{equation}

Once the update term is less than a preset threshold, the solution can be considered to converge.
\begin{equation}
    \label{equation6}
    \begin{gathered}
        \check{\mathbf{x}}_{i+1} \Leftarrow \hat{\mathbf{x}}_{i+1}^{j+1}, \; \check{\mathbf{P}}_{i+1} \Leftarrow ( \mathbf{I} -  \mathbf{K}\mathbf{H} )\hat{\mathbf{P}}_{i+1}
    \end{gathered}
\end{equation}

With the iterative update of ESKF, the AC-LIO system could further promote the convergence of the state estimation via the asymptotic distortion compensation and convergence criterion introduced in following sections.

\subsection{Asymptotic Distortion Compensation} \label{asymptotic compensation}

Compared {with LiDAR measurement noise}, motion distortion may introduce bias in pointcloud registration, making it difficult to utilize the statistical properties of pointcloud to mitigate the adverse effects on state estimation.
During the on-manifold propagation of state $\mathbf{x}$, the {covariance} of current error state $\delta \hat{\mathbf{x}}_{i+1}$ consists of the propagation of previous state $\delta \check{\mathbf{x}}_{i}$ and the accumulated process noise. The covariance between adjacent error states can be expressed as follows:
\begin{equation}
    \label{equation7}
    \begin{gathered}
            \delta \check{\mathbf{x}}_i \sim \mathcal{N} (\mathbf{0}_{18\times 1}, \check{\mathbf{P}}_{i}) \\
            \delta \hat{\mathbf{x}}_{i+1} \sim \mathcal{N} (\mathbf{0}_{18\times 1}, \mathbf{F_p} \check{\mathbf{P}}_{i} \mathbf{F_p}^T + \mathbf{F_n} \mathbf{Q} \mathbf{F_n}^T)
    \end{gathered}
\end{equation}

With reference to the Rauch–Tung–Striebel fixed-interval smoothing strategy \cite{gelb1974applied}, the update term $\delta \mathbf{x}$ can be backpropagated along the prior state chain.
The contribution of previous state in current update term just corresponds to the proportion of propagation covariance $\mathbf{F_p} \check{\mathbf{P}}_{i} \mathbf{F_p}^T$ within current covariance $\hat{\mathbf{P}}_{i+1}$. Then, the gain matrix $\mathbf{G}$ and the backward recursion of $\delta \mathbf{x}$ can be denoted as follows:
\begin{equation}
    \label{equation8}
    \begin{gathered}
        \mathbf{G} = \check{\mathbf{P}}_{i} \mathbf{F_p}^T \hat{\mathbf{P}}_{i+1}^{-1} \\
        \delta \hat{\mathbf{x}}_{i}^{j} = \mathbf{G} \cdot \delta \hat{\mathbf{x}}_{i+1}^{j}
    \end{gathered}
\end{equation}

Furthermore, this sequential backpropagation method can be extended to any states within the LiDAR frame, enabling selective updates of several anchors in the prior state chain. This approach can reduce the computational complexity of both backpropagation and distortion compensation.
For any two states {$\mathbf{x}_i$ and $\mathbf{x}_{i+n}$} in the prior state chain, the propagation matrix between them is denoted as $\mathbf{F_p}^{i|i+n}$, and the corresponding backward gain matrix is denoted as $\mathbf{G}_{i}^{i+n}$. Then the backpropagation relationship can be expressed as:
\begin{equation}
    \label{equation9}
    \begin{gathered}
        \mathbf{F_p}^{i|i+n} = \prod_{k=i}^{i+n-1} \mathbf{F_p}^{k|k+1} \\
        \mathbf{G}_{i}^{i+n} = \check{\mathbf{P}}_{i} ( \mathbf{F_p}^{i|i+n} )^T ( \hat{\mathbf{P}}_{i+n} )^{-1} \\
        \check{\mathbf{x}}_{i}^{j} = \check{\mathbf{x}}_{i} \boxplus ( \mathbf{G}_{i}^{i+n} \cdot \delta \hat{\mathbf{x}}_{i+n}^{j} ) 
    \end{gathered}
\end{equation}

After each backpropagation of the update term $\hat{\mathbf{x}}_{i+n}^{j}$, where the $j$ {denotes} the current iteration times, the residual pointcloud distortion can {be further} compensated based on the smoothed trajectory. 
For the point $\mathbf{p}^L$ in LiDAR frame, and the state corresponding to its timestamp $t$ after the $j$th backpropagation can be denoted as $\hat{\mathbf{x}}_{t}^{j}$, then the updated coordinates can be noted as:
\begin{equation}
    \label{equation10}
    \begin{gathered}
        \hat{\mathbf{p}}_{j}^{G} = \mathbf{T}(\hat{\mathbf{x}}_{t}^{j}) \mathbf{T}_L^I \mathbf{p}^L
    \end{gathered}
\end{equation}

Where $\mathbf{T}(\mathbf{x})$ represents the transformation matrix of state $\mathbf{x}$ in frame $G$. As the iterative ESKF converges, the final solution of the backward propagation can be expressed as:
\begin{equation}
    \label{equation11}
    \begin{gathered}
        \bar{\mathbf{x}}_{i} \Leftarrow \check{\mathbf{x}}_{i}^{j+1}, \; \bar{\mathbf{P}}_{i} \Leftarrow \mathbf{G}_{i}^{i+n} \check{\mathbf{P}}_{i+n}(\mathbf{G}_{i}^{i+n})^T
    \end{gathered}
\end{equation}

In practice, if there is no need to reuse the previous state queue in next frame, the covariance update after backpropagation can be skipped for efficiency.

\subsection{Convergence criterion} \label{convergence criterion}

Due to the lack of correlation between the previous error and the current motion distortion within a LiDAR frame, it is necessary to introduce a convergence criterion to regulate the asymptotic distortion compensation.
If the previous LiDAR frame has sufficiently converged, the current frame may start with a relatively small initial error, allowing the backpropagation to effectively compensate for motion distortion during iterative state estimation.
However, if {the} previous LiDAR frame has not converged well, then the initial error may degrade the system accuracy via the backpropagation.
Therefore, a simple and efficient method is required to detect the convergence status of each LiDAR frame, in order to guide the asymptotic distortion compensation.

\begin{figure}[t!]
    \vspace{2mm}
    \begin{center}
        {\includegraphics[width=0.9\columnwidth]{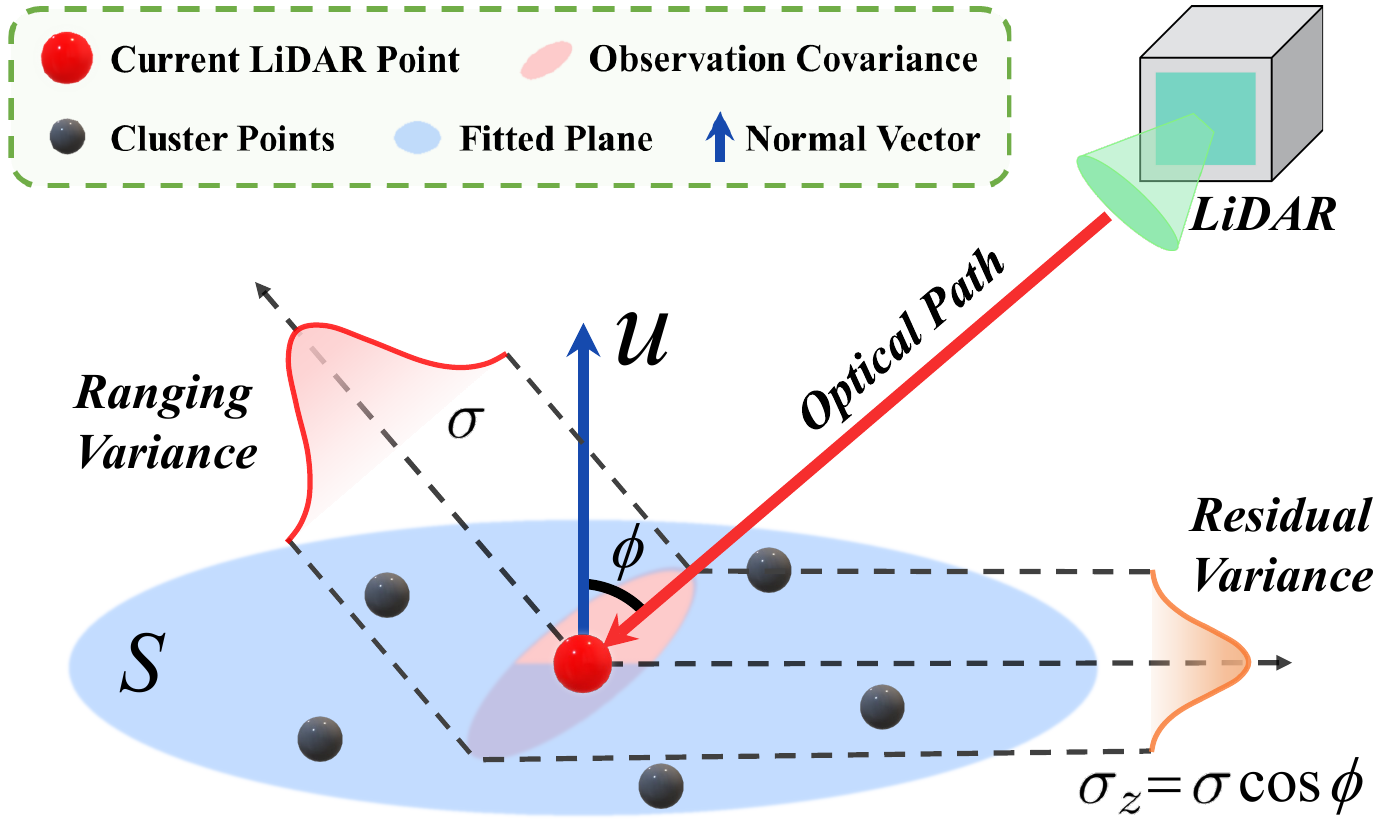}}
    \end{center}
    \vspace{-0.3cm}
    \caption{\label{fig:covariance}The relationship between original ranging variance of LiDAR point and point-to-plane residual variance, where the residual $z$ is a scalar.}
    \vspace{-2mm}
\end{figure}

To exploit the statistical properties of the pointcloud, during iterative state estimation, the mean value of the point-to-plane residual (abbreviated as APR above) is selected as the criterion for evaluating the convergence of pointcloud registration.
For simplification, assuming the pointcloud registration converges to the true pose and only considering the ranging accuracy of LiDAR, the expectation of the APR can be modeled to provide a threshold reference for the convergence criterion.
Denote the incident angle of point $\mathbf{p}$ to the fitted plane $S$ as $\phi$, which can be expressed as $\cos^{-1} \tfrac{|\mathbf{u} \cdot \hat{\mathbf{p}}^G|}{|\mathbf{u}| \cdot |\hat{\mathbf{p}}^G|}$ according to Eq.~\ref{equation3}. 
Here, $\mathbf{z}$ is the point-to-plane residual vector and $z$ is its scalar value. The ranging error of a LiDAR point is assumed to follow a Gaussian distribution with zero mean and standard deviation $\sigma$.
When the laser beam is perpendicular to the fitted plane, {the normalized squared residual $\tfrac{{z}_{\perp}^2}{\sigma^2}$ follows} a chi-squared distribution and can be expressed as:
\begin{equation}
    \label{equation12}
    \begin{gathered}
        {z}_{\perp} = |\mathbf{p}^{G} - \hat{\mathbf{p}}^{G}| \sim \mathcal{N}(0, \sigma^2) \\
        \frac{{z}_{\perp}^2}{\sigma^2} \sim \chi^{2} (1), \; \mathbb{E}({z}_{\perp}) = \sigma
    \end{gathered}
\end{equation}

As the surroundings change during the system's ego-motion, the distribution of the normal vector $\mathbf{u}$ to the plane $S$ also varies.  
This variation in {the} geometrical scene is reflected in the distribution of the incident angle $\phi$.  
From Fig.~\ref{fig:covariance}, the standard deviation of this residual $\sigma_z$ is given by $\sigma\cos\phi$.
Assuming that the incident angle is uniformly distributed over the domain $[0, \pi/2)$, the expected value of the point-to-plane residual can be expressed as:
\begin{equation}
    \label{equation13}
    \begin{gathered}
        \mathbb{E}({z}) = \frac{2}{\pi} \int_{0}^{\frac{\pi}{2}} \mathbb{E}({z}_{\perp}) \cos{\phi} \; d\phi = \frac{2}{\pi} \sigma
    \end{gathered}
\end{equation}

Therefore, assuming the pointcloud registration converges to the true pose, and only considering the LiDAR ranging error, the expectation of APR should be ${2\sigma}/{\pi}$.
\hl{This threshold is used to determine the convergence degree of pointcloud registration, and to control the asymptotic distortion compensation.}
Noting that $\sigma_t=\mathbb{E}({z})$ and the selected threshold of convergence criterion as $\eta$, due to the factors such as dynamic objects and cumulative errors in pointcloud map, the threshold $\eta$ can be appropriately extended to the interval of $[\sigma_t, 2\sigma_t]$ in practice.
\hl{The uniform incident-angle assumption is a simplification, and structured indoor scenes and unstructured outdoor scenes may exhibit different incident-angle distributions.}
A static-movement-static sequence was collected with Livox AVIA to evaluate the validity of {the} proposed convergence criterion.
As shown in Fig.~\ref{fig:APR}, the value of $\sigma_t$ is very close to the APR during the static phases.

\hl{During iterative state estimation, two conditions are checked.}
First, the previous LiDAR frame is evaluated: if its APR is less than the threshold $\eta$, it is considered to be sufficiently converged and the process continues. 
Second, the convergence of the current registration is checked: if the APR still remains below $\eta$, there is no need to backpropagate for small residual distortion; 
otherwise, the backpropagation and compensation are performed during the iteration to correct the residual distortion.
On one hand, this approach can ensure {that} the backpropagation and asymptotic compensation {are} performed {only when} the previous LiDAR frame {has converged sufficiently}, thus enhancing the odometry robustness in long-term and large-scale scenario. On the other hand, it can {avoid additional} operations in unreliable or unnecessary conditions, to improve the system computational efficiency. 

\section{EXPERIMENT RESULTS}

In this section, the AC-LIO framework is qualitatively and quantitatively evaluated through a series of experiments with various public and self-collected datasets.
To examine its performance in long-term and large-scale scenarios, extensive experiments are conducted on the odometry accuracy, the validity of the convergence criterion, and the time efficiency of the system.

\subsection{Experiment Dataset}

The datasets cover a wide range of structured environments (e.g., campus road and garage) and unstructured settings (e.g., botanical gardens), as well as different {platforms} (helmet, vehicle, handheld) to boost the challenge and completeness of assessment.

\begin{figure}[t]
    \vspace{1.5mm}
    \begin{center}
        {\includegraphics[width=0.9\columnwidth]
        {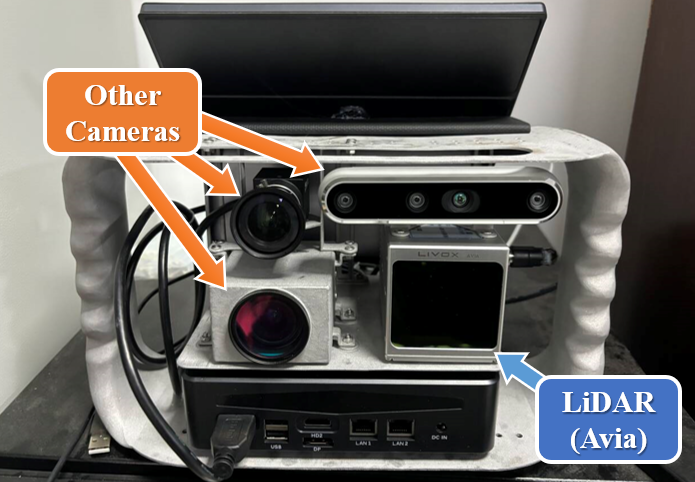}}
    \end{center}
    \vspace{-0.2cm}
    \caption{\label{fig:Sensors}The handheld sensor suite for self-collected {datasets}, including solid-state LiDAR Livox Avia with built-in BMI088 IMU, and three other cameras.}
    \vspace{-2mm}
\end{figure}

\begin{itemize}

\item WHU-Helmet~\cite{li2023whu}: Collected {with a} compact helmet system with Livox LiDAR, the dataset contains complex large-scale scenes including wild, urban{, and} tunnel, with challenges of drastic motion from wearer and repeated symmetrical geometric structures.

\item BotanicGarden~\cite{liu2024botanicgarden}: Mounted on an all-terrain wheeled robot, this dataset consists of diverse natural scenes including riversides, narrow trails, and bridges, \hl{with LiDAR measurements from Livox Avia and VLP16, and IMU measurements from BMI088 and MTi-680G with different noise levels.}

\item {MCD~\cite{nguyen2024mcd} : Collected across three campus environments with multiple sensors including spinning and semi-solid LiDARs, IMUs, and cameras, this dataset features diverse and challenging scenarios with high-accuracy ground truth and large-scale mapping.}

\item Handheld datasets: \hl{Multiple datasets under different scenarios and motion conditions were collected using Livox Avia, covering indoor and outdoor scenes such as a library, campus roads, and a UAV airfield. Strictly closed-loop trajectories were also conducted for end-to-end error evaluation.} The {handheld device} is shown in Fig.~\ref{fig:Sensors}.

\end{itemize}

\begin{table*}[t]
\vspace{2mm}
\footnotesize
\centering
\renewcommand{\arraystretch}{1.1}
\vspace{1mm}
\caption{Odometry Accuracy Evaluation - RMSE of ATE \& End to End Errors($m$)}
\vspace{-1mm}
\label{tab:Accuracy Evaluation}
\begin{threeparttable}

\resizebox{1.0\textwidth}{!}{

\begin{tabularx}{0.95\linewidth}{@{}P{1.3cm}P{1.6cm}H{2.8cm}CCCCC@{}}
\toprule
\textbf{Dataset} & \textbf{LiDAR Type} & \textbf{Sequence} & \textbf{Ours} & \textbf{FAST-LIO2} & \textbf{DLIO} & \textbf{SLICT} & \textbf{Point-LIO}  \\
\midrule

\multirow{4}{*}{\begin{tabular}[c]{@{}c@{}} \textbf{WHU}\tnote{1} \\ \textbf{Helmet} \end{tabular}}
& \multirow{4}{*}{\begin{tabular}[c]{@{}l@{}} {Livox} \\ {AVIA} \end{tabular}} 
& {Residence\_5.2 (452.9m)}
  & \textbf{0.415} & \underline{0.505} & 4.660 & 1.550 & $\times$ \\

&
& {Parking\_3.2 (628.9m)}
  & \textbf{0.582} & 0.618 & 7.725 & \underline{0.602} & $\times$ \\

&
& {Subway\_3.3 (827.0m)}
  & \textbf{1.670} & 3.752 & 3.688 & \underline{1.967} & $\times$ \\

&
& {Mall\_6.3 (1006.5m)}
  & \underline{0.216} & \textbf{0.208} & 3.238 & 0.274 & 15.409 \\

\noalign{\vskip 0.5mm}
\hline
\noalign{\vskip 0.8mm}

\multirow{6}{*}{\begin{tabular}[c]{@{}c@{}} \textbf{Botanic}\tnote{1} \\ \textbf{Garden} \end{tabular}}
& \multirow{3}{*}{\begin{tabular}[c]{@{}l@{}} {Livox} \\ {AVIA} \end{tabular}}
& {Garden\_1005 (566.8m)}
  & \underline{0.773} & 1.053 & \textbf{0.640} & 7.931 & 0.917 \\

&
& {Garden\_1006 (686.2m)}
  & \textbf{1.796} & \underline{2.075} & 3.017 & 6.527 & 2.471 \\

&
& {Garden\_1008 (855.9m)}
  & \textbf{1.152} & \underline{1.577} & 14.829 & 4.445 & 1.590 \\

\noalign{\vskip 0.4mm}
\cline{2-8}
\noalign{\vskip 0.8mm}

& \multirow{3}{*}{{VLP16}} 
& {Garden\_1005 (566.8m)}
  & \textbf{0.175} & 0.244 & 0.565 & 37.425 & 0.750 \\

&
& {Garden\_1006 (686.2m)}
  & \textbf{0.377} & \underline{0.388} & 0.393 & $\times$ & 0.445  \\

&
& {Garden\_1008 (855.9m)}
  & \textbf{0.339} & \underline{0.363} & 0.694 & $\times$ &  0.457 \\
  
\noalign{\vskip 0.5mm}
\hline
\noalign{\vskip 0.8mm}

\multirow{4}{*}{\textbf{MCD}\tnote{1}}
& \multirow{4}{*}{\begin{tabular}[c]{@{}l@{}} \hl{Livox} \\ \hl{Mid-70} \end{tabular}} 
& {ntu\_day\_01 (3197.7m)}
  & \textbf{0.598} & \underline{0.731} & 1.925 & 1.890 & 28.858 \\

&
& {ntu\_day\_02 (641.8m)}
  & \underline{0.324} & 0.354 & 0.636 & \textbf{0.168} & 0.391 \\

&
& {ntu\_night\_04 (1459.1m)}
  & \textbf{0.639} & \underline{0.708} & 2.373 & 1.002 & 14.145 \\

&
& {ntu\_night\_08 (2420.7m)}
  & \textbf{0.804} & 1.204 & 2.056 & \underline{0.822} & $\times$ \\





\noalign{\vskip 0.5mm}
\hline
\noalign{\vskip 0.8mm}

\multirow{6}{*}{\begin{tabular}[c]{@{}c@{}} \textbf{Handheld}\tnote{2} \end{tabular}}
& \multirow{6}{*}{\begin{tabular}[c]{@{}l@{}} {Livox} \\ {AVIA} \end{tabular}} 
& {Indoor\_1 (\textasciitilde 100m)}
  & \textbf{0.893} & \underline{1.133} &  $\times$ & 1.960 & 5.401 \\

&
& {Indoor\_2 (\textasciitilde 100m)}
  & 0.227 & \underline{0.204} & 3.354  & 0.249 & \textbf{0.164} \\

&
& {Indoor\_3 (\textasciitilde 100m)}
  & \textbf{1.596} & 3.001 & $\times$ & \underline{2.476} & 6.844 \\

&
& {Outdoor\_1 (\textasciitilde 300m)}
  & \textbf{0.138} & 1.901 & \underline{0.486} & 1.835 & 0.728 \\

&
& {Outdoor\_2 (\textasciitilde 400m)}
  & \textbf{0.984} & 2.194 & 7.879 & 1.803 & \underline{1.261} \\

&
& {Outdoor\_3 (\textasciitilde 500m)}
  & \underline{2.666} & 4.702 & \textbf{1.445} & 4.127 & 4.607 \\

\bottomrule
\noalign{\vskip 1mm}
\textbf{Average(RMSE)} & & \hl{(Handheld excluded)} & \textbf{0.704} & \underline{0.984} & 3.317 & 5.384 & 6.543 \\
\noalign{\vskip 0mm}
\bottomrule
\end{tabularx}}
\begin{tablenotes}
\footnotesize
\item ~$^1$ ~means the dataset is evaluated by RMSE of ATE, and ~$^2$ means it is evaluated by End to End errors.
\item $\times$ ~denotes the system severely diverged midway,  and $\underline{\hspace{0.25cm}}$ ~denotes the {second-best} result compared to bolded result.
\end{tablenotes}
\end{threeparttable}
    \vspace{-2mm}
\end{table*}

\begin{figure*}[thbp]
    \vspace{0.5mm}
    \begin{center}
        {\includegraphics[width=2.0\columnwidth]
        {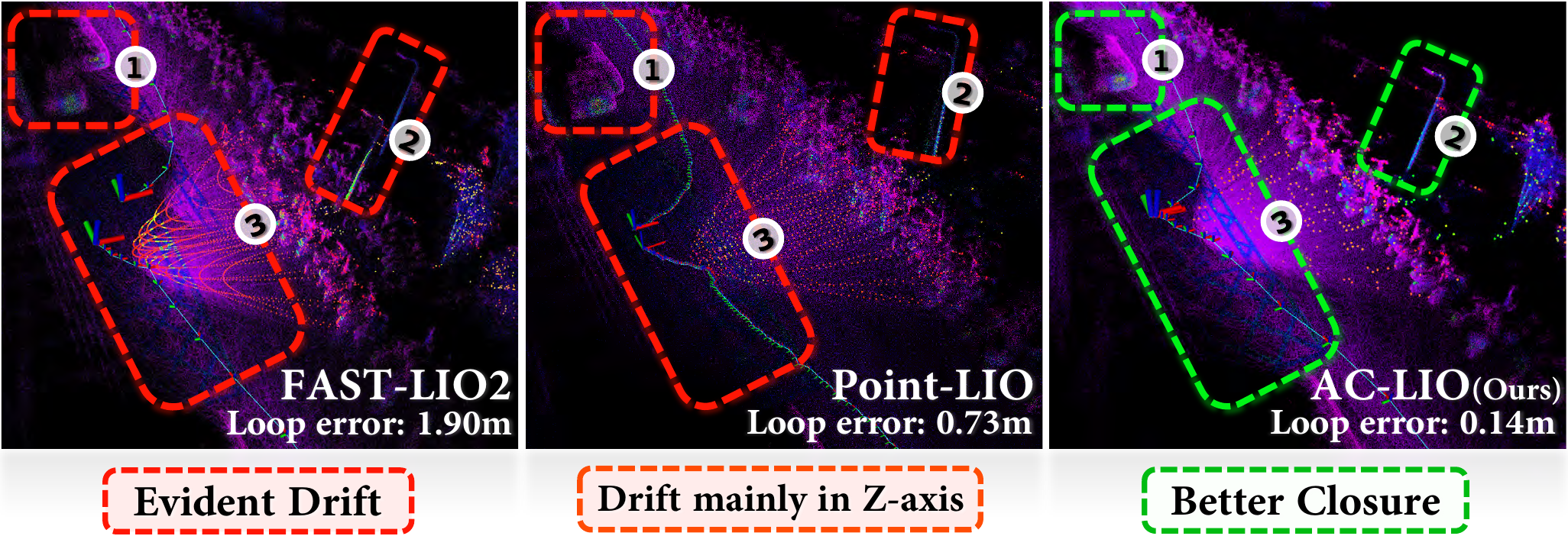}}
    \end{center}
    \vspace{-3mm}
    \caption{\label{fig:Acc Evaluation}\textbf{Accuracy evaluation in campus road scene.} The figure shows the result of sequence Outdoor\_1. Within the dashed box are, in order, the car, the streetlight, and the pattern on the ground. After initialization, this sequence was collected {while running} to enhance the motion intensity and complete a closed loop of about 300m. The closure trajectory and the map consistency suggest that {the} AC-LIO system achieves better performance.}
    \vspace{-4mm}
\end{figure*}

\subsection{Experiment Settings} 

\begin{itemize}
    \item The AC-LIO was compared with current state-of-the-art LIO frameworks \hl{for motion distortion mitigation}, rather than LO frameworks without IMU integration. \hl{Some frameworks designed for spinning LiDAR were not included in comparison.} Among the LIO frameworks, \textbf{FAST-LIO2} compensates for distortion via IMU prior trajectory before iterative estimation, \textbf{DLIO} applies a coarse-to-fine pointcloud deskewing approach, \textbf{SLICT} utilizes continuous-time state estimation to optimize the motion distortion, and \textbf{Point-LIO} is an extremely {high-frequency} LIO with a point-by-point strategy to avoid motion distortion, \hl{showing stronger robustness to IMU saturation than FAST-LIO2 and AC-LIO.}

    \item All algorithms are implemented in C++ and employed on an i9 CPU computer with the ROS in Ubuntu. These methods only utilize CPU for computation without the assistance of GPU parallel computing. 
    As the AC-LIO is implemented on the basis of FAST-LIO2, the parameters such as imu\_noise, voxel\_size, max\_iteration, and point\_filter\_num are kept consistent especially with FAST-LIO2 for a fair comparison. The parameters of other methods are also carefully fine-tuned to achieve their best performance.

    \item Since the AC-LIO computes the convergence criterion to guide the asymptotic distortion compensation during iterative update, the number of iterations in each optimization epoch is fixed (i.e., for each LiDAR frame).
    In contrast, FAST-LIO2 {terminates} the iteration as the update gradient {falls below} a threshold, which may lead to an accuracy difference with AC-LIO. 
    Therefore, two groups of ablation studies were conducted (Sec.~\ref{criterion evaluation}) specifically to validate the performance of selective intra-frame smoothing.
\end{itemize}

\subsection{Odometry Accuracy Evaluation}

\begin{figure*}[thbp]
    \vspace{2mm}
    \begin{center}
        {\includegraphics[width=2\columnwidth]
        {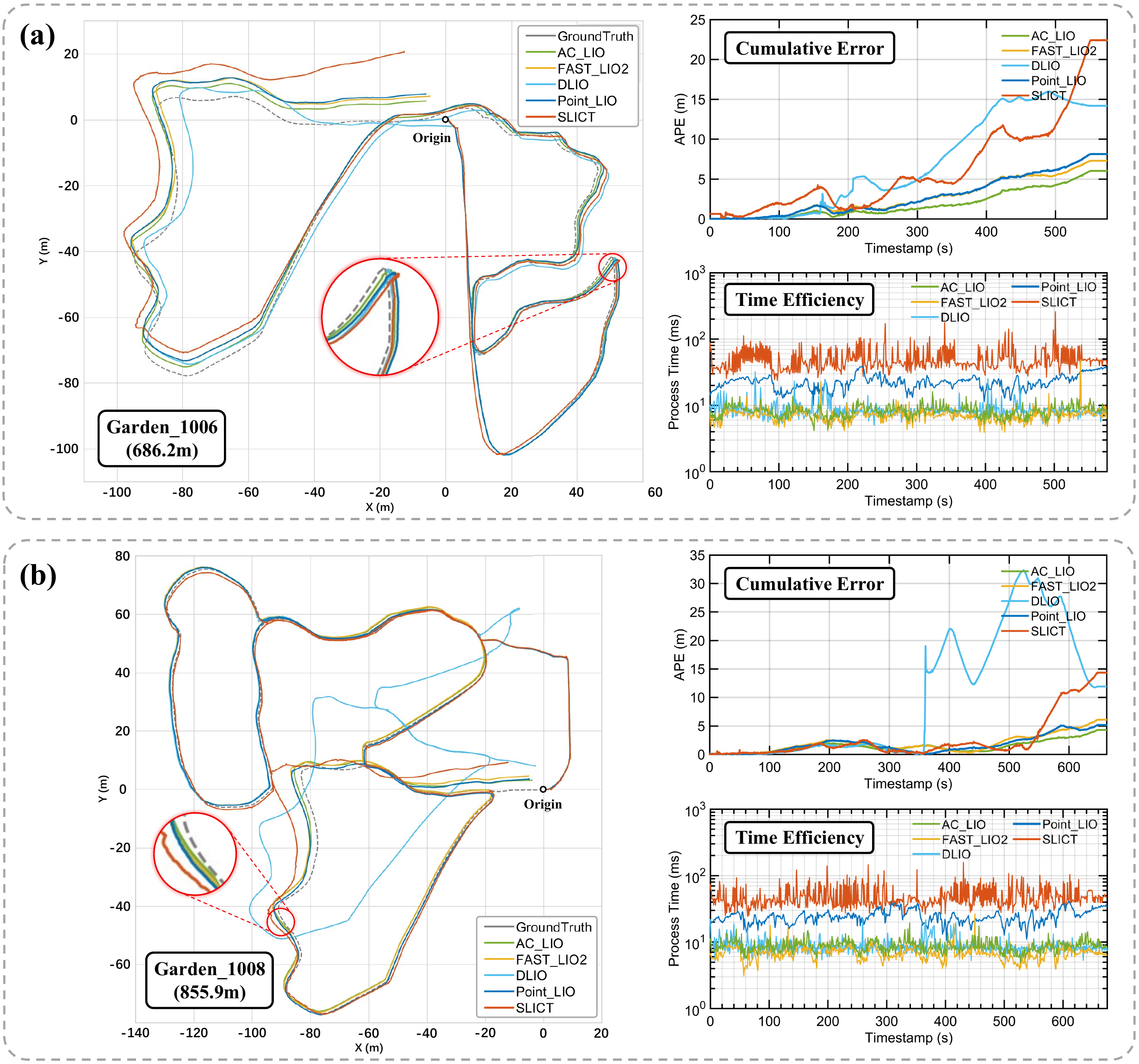}}
    \end{center}
    \vspace{-3.5mm}
    \caption{\label{fig:Acc Evaluation Garden}\textbf{Accuracy evaluation in Botanic Garden dataset.} The figure shows the result of sequence \textbf{(a) Garden\_1006} and \textbf{(b) Garden\_1008}. For each sequence, the left part presents the trajectory comparison, with the top 10\% states aligned to the ground truth, and the right part displays the cumulative error curve of ATE (m) along with the corresponding real-time process efficiency. The results suggest that AC-LIO system achieves lower cumulative error while maintaining relatively good time efficiency (Sec.~\ref{time efficiency}). The average runtime of {AC-LIO} is only slightly higher than the FAST-LIO2 framework.}
    \vspace{-4mm}
\end{figure*}

The odometry accuracy evaluation aims to demonstrate the overall accuracy performance of the selected LIO frameworks and AC-LIO.
In the evaluation of WHU-Helmet, the sequences $\textit{Residence\_5.2}$ and $\textit{Subway\_3.3}$ involve aggressive motion combined with locally degraded scenarios, while $\textit{Parking\_3.2}$ and $\textit{Mall\_6.3}$ are affected by extensive {pedestrian occlusions}. 
These factors impose higher demands on the robustness of the LIO system.
For BotanicGarden {dataset}, both the solid-state (Livox AVIA) and the spinning (VLP16) LiDAR were tested. 
{The smooth motion of the wheeled robotic platform makes the dataset suitable for evaluating long-term drift under regular motion conditions.}
However, the abundance of unstructured scenes in the garden (e.g., thick woods, bushes) presents challenges for constructing {geometric} constraints.
{On MCD dataset, AC-LIO demonstrates superior performance in typical campus scenarios. The results of SLICT are quoted from the MCD paper.}
In self-collected datasets, \hl{the sensor suite traversed multiple closed-loop paths across diverse scenarios, and the localization accuracy was evaluated by the end-to-end error.}
In $\textit{Indoor\_1}$ UAV airfield and $\textit{Indoor\_3}$ library scenarios, intense turning was performed in narrow environments. 
\hl{The $\textit{Outdoor}$ sequences cover different motion patterns, including walking, running, and sharp turns on campus roads.}

Concrete experiment results are demonstrated in Table~\ref{tab:Accuracy Evaluation}, Fig.~\ref{fig:Acc Evaluation} and Fig.~\ref{fig:Acc Evaluation Garden}. {This approach} further optimizes the residual motion distortion via selective intra-frame smoothing strategy, thus improving the accuracy while ensuring the \hl{system robustness} under challenging scenarios. 
The results \hl{show that AC-LIO} generally outperforms other benchmark frameworks in terms of accuracy during long-term \hl{and} large-scale scenes, {achieving a 28.5\% relative reduction in average RMSE across sequences compared with the second-best method, while maintaining} high robustness under challenging scenarios.

\subsection{Convergence criterion Evaluation} \label{criterion evaluation}

\begin{table*}[t]
\vspace{2mm}
\footnotesize
\centering
\renewcommand{\arraystretch}{1.0}
\caption{{Validity Evaluation of the Convergence Criterion} - RMSE of ATE ($m$)}
\label{tab:criterion RMSE}
\begin{threeparttable}

\resizebox{1.0\textwidth}{!}{

\begin{tabularx}{0.95\linewidth}{@{\hspace{1mm}}H{1.5cm}C|P{2.5cm}|CCCCCCC@{\hspace{1mm}}}
\toprule
\textbf{Sequence} & \textbf{LiDAR} & \textbf{\hl{FAST-LIO2\,(ori\,/\,fix)}} & 0${\sigma_t}$ & 0.5$\sigma_t$ & 1$\sigma_t$ & 1.5$\sigma_t$ & 2$\sigma_t$ & 2.5$\sigma_t$ & 3$\sigma_t$  \\
\midrule

\multirow{2}{*}{\begin{tabular}[c]{@{}l@{}} 
{Garden\_1005} \\ (566.8m)
\end{tabular} } 
& AVIA & 1.053\,/\,\hl{1.015} & 1.063 & 1.063 & 1.004 & 0.943 & \textbf{0.743} & \underline{ 0.907} & 1.120 \\

\noalign{\vskip 0.4mm}
\cline{2-10}
\noalign{\vskip 0.8mm}

& VLP16 & 0.244\,/\,\hl{0.222} & 0.244 & 0.241 & 0.226 & \textbf{0.175} & \underline{0.193} & 0.208 & 0.233 \\

\bottomrule
\end{tabularx}}
\end{threeparttable}
    \vspace{0.0cm}
\end{table*}

\begin{figure}[thbp]
    \vspace{-1mm}
    \begin{center}
        {\includegraphics[width=1.0\columnwidth]
        {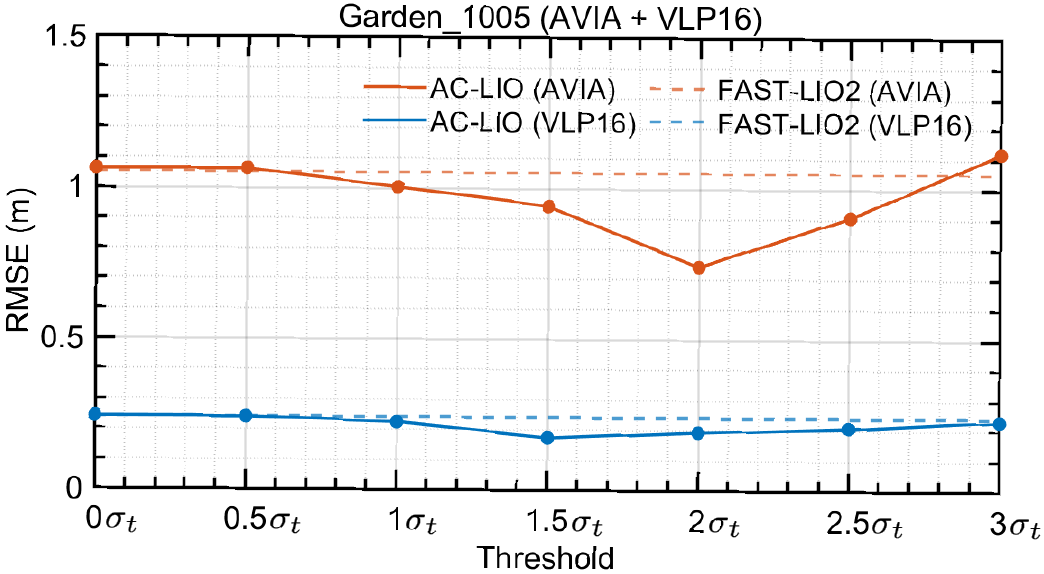}}
    \end{center}
    \vspace{-5mm}
    \caption{\label{fig:Convergence Curve}\textbf{Validity evaluation of convergence criterion.} The figure presents the ablation experiments on convergence criterion using Garden\_1005 sequence: \textbf{(1) Orange curves for Livox AVIA and (2) Blue curves for VLP16}, demonstrating the generalization across different LiDAR sensors. The dashed lines \hl{represent the RMSE of FAST-LIO2(ori).} As threshold $\eta$ increases from zero, the accuracy improves because more sufficiently converged states are allowed to perform backpropagation. However, as $\eta$ continues to increase, states with larger initial errors are also allowed to backpropagate, which gradually degrades the overall accuracy.}
    \vspace{-4mm}
\end{figure}

\begin{figure}[t]
    \vspace{1.5mm}
    \begin{center}
    {\includegraphics[width=0.98\columnwidth]
        {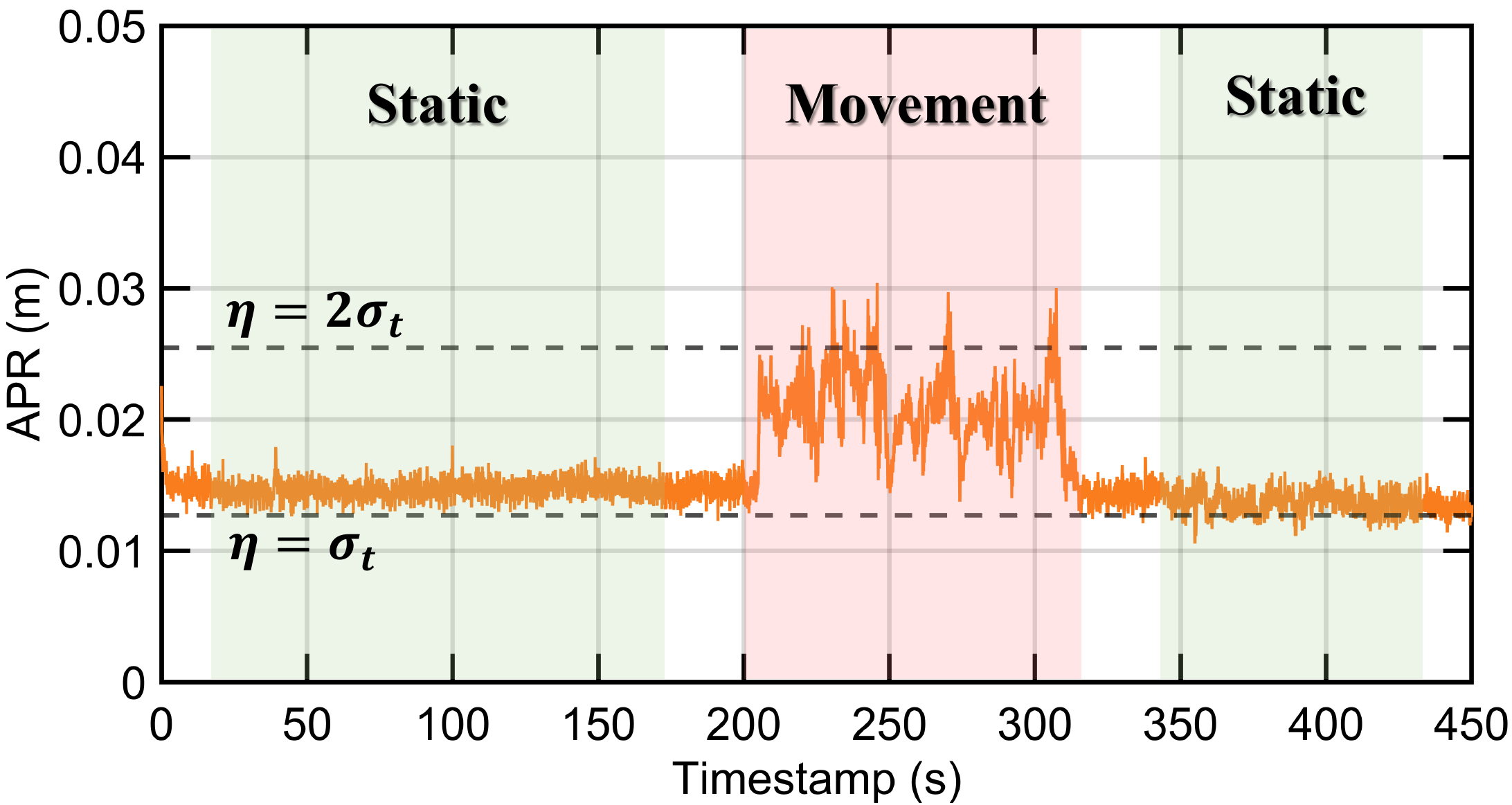}}
    \end{center}
    \vspace{-4mm}
    \caption{\label{fig:APR}\textbf{Rationality evaluation of convergence criterion.} The orange curve {refers to} the average point-to-plane residual (APR) during a static-movement-static sequence in an indoor scenario. The APR values under static phases are very {close} to the theoretical value of $\sigma_t$, indicating good consistency between {the} theoretical derivation and actual APR.}
    \vspace{-0.4cm}
\end{figure}

To evaluate the rationality and validity of the proposed convergence criterion, two types of experiments were conducted.\\
\noindent \textbf{Rationality}: Since the convergence criterion is based on certain simplifications and assumptions, the rationality was examined by comparing the actual APR values with the theoretical value of $\sigma_t$, as the APR during the static phase can be regarded as a reference for $\sigma_t$ under ideal conditions.
A static-movement-static sequence was collected with Livox AVIA. 
As shown in Fig.~\ref{fig:APR}, the actual APR values during the static phases were very close to $\sigma_t$, indicating good consistency between theoretical derivation of $\sigma_t$ and the actual APR results.
Moreover, the APR increased and became more volatile during the movement. 
Too large an APR reflects a higher risk of insufficient convergence. 
The convergence criterion is introduced to prevent asymptotic distortion compensation from starting with too high an initial error.

\noindent \textbf{Validity}: This ablation study evaluates the accuracy of AC-LIO under different $\eta$ threshold levels for both AVIA and VLP16, with FAST-LIO2 serving as the benchmark.
While keeping other parameters the same, only the threshold of the convergence criterion was adjusted.
The experimental results are presented in Table~\ref{tab:criterion RMSE} and Fig.~\ref{fig:Convergence Curve}.

\begin{itemize}

\item $\eta \in [0, 0.5\sigma_t]$: No or rare backpropagation and asymptotic compensation is triggered under this condition, reflecting the basic odometry accuracy.

\item $\eta \in [\sigma_t, 2\sigma_t]$: Recommended interval in consideration of other error factors, assuring {that} backpropagation and asymptotic compensation are only performed with sufficient convergence in previous frame.

\item $\eta \in [2.5\sigma_t, 3\sigma_t]$: Too high a threshold may cause the backpropagation to operate on states with large initial errors, potentially degrading the overall system accuracy and robustness.

\end{itemize}

The results show that selecting the threshold $\eta$ within the range of $[\sigma_t, 2\sigma_t]$ tends to yield higher odometry accuracy.
\hl{For a fair comparison, both the original FAST-LIO2\,(ori) and a fixed-iteration variant are reported, where FAST-LIO2\,(fix) uses the same fixed number of iterations as AC-LIO.}
As the threshold $\eta$ increases from zero, the odometry accuracy initially improves because more sufficiently converged states are allowed to perform backpropagation and asymptotic compensation. 
However, as $\eta$ continues to increase, states with larger initial errors are also allowed to backpropagate, which diminishes the effectiveness of asymptotic compensation and gradually degrades odometry accuracy.

\begin{table}[t]
\footnotesize
\centering
\vspace{1mm}
\renewcommand{\arraystretch}{1.1}
\caption{Time Consumption Evaluation ($ms$)}
\vspace{-1mm}
\label{tab:Time Consumption}
\begin{threeparttable}

\begin{tabularx}{1\linewidth}{@{}H{1.4cm}CCCCC@{}}
\toprule
& \textbf{Ours} & \textbf{FAST-LIO2} & \textbf{DLIO} & \textbf{SLICT} & \textbf{Point-LIO}  \\
\midrule
\textbf{~\circnum{1}~Min} & 4.12 & \textbf{2.89} & \underline{3.95} & 22.61 & 7.83 \\
\textbf{~\circnum{2}~Max} & 70.69 & 66.38 & \textbf{37.22} & 460.24 & \underline{50.84} \\
\textbf{~\circnum{3}~Mean} & \underline{8.55} & \textbf{7.18} & 8.57 & 48.62 & 24.48 \\
\textbf{~\circnum{4}~Median} & 8.32 & \textbf{6.97} & \underline{8.24} & 42.91 & 23.61 \\
\bottomrule
\end{tabularx}
\end{threeparttable}
    \vspace{-2mm}
\end{table}

\begin{figure}[t]
    \vspace{-1mm}
    \begin{center}
        {\includegraphics[width=0.98\columnwidth]
        {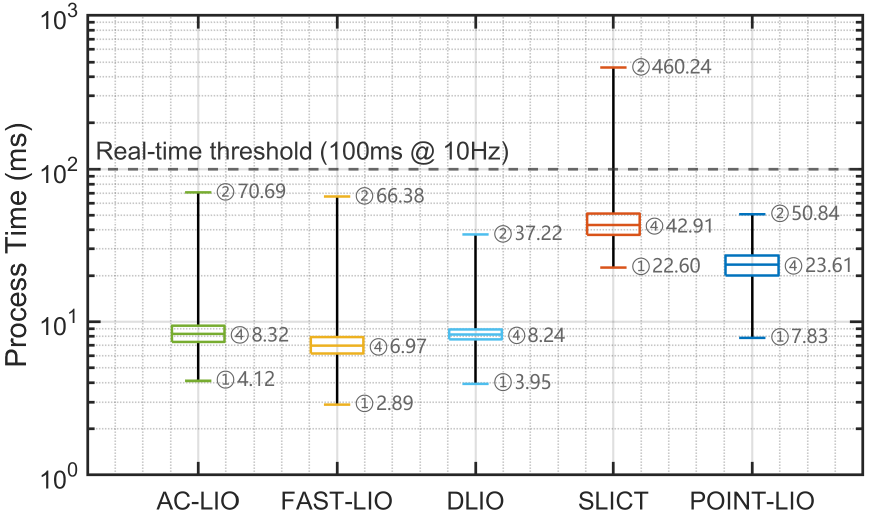}}
    \end{center}
    \vspace{-4mm}
    \caption{\label{fig:time_cost}\textbf{Time efficiency evaluation.} {The lower and upper whiskers correspond to} \textbf{\circnum{1}~{Min}} {and} \textbf{\circnum{2}~{Max}}{, respectively.} The cross line inside box denotes the \textbf{\circnum{4}~{Median}} {processing time}. {AC-LIO} achieves the second-highest efficiency, only behind {FAST-LIO2} and comparable to the lightweight DLIO. {Moreover, }the maximum runtime also meets the real-time requirements.}
    \vspace{-4mm}
\end{figure}

\subsection{Time Efficiency Evaluation} \label{time efficiency}

The runtime of each algorithm during iteration was evaluated, with the results summarized in Table~\ref{tab:Time Consumption} and Fig.~\ref{fig:time_cost}. 
The corresponding runtime curves are presented in Fig.~\ref{fig:Convergence Curve}.  
In terms of average processing time, the AC-LIO framework ranks second among the tested frameworks, just following the FAST-LIO2 framework and comparable to the lightweight DLIO framework. 
The maximum iteration time also meets the realtime {demand} (below 100\,ms for a 10\,Hz LiDAR input).  
Owing to the convergence criterion strategy, AC-LIO avoids incurring additional optimization overhead under unreliable or unnecessary conditions, thereby maintaining a balance between system convergence and high time efficiency.

   
\section{Conclusion}

{This work proposes AC-LIO, a novel discrete-state} LIO framework with selective intra-frame smoothing.
{This work further examines the factors} that may introduce errors in long-duration, large-scale scenarios, aiming to further enhance the convergence of LIO system.
The approach of guiding asymptotic compensation based on convergence criterion is designed to further suppress the motion distortion within LiDAR frame, improving the odometry accuracy while ensuring system robustness.
Extensive experiments indicate that AC-LIO framework achieves an average RMSE reduction of approximately {28.5\%} over the {second-best} method, demonstrating higher accuracy in long-term, large-scale localization and mapping. 
In terms of time efficiency, {AC-LIO} framework ranks second place among the evaluated frameworks, just following the FAST-LIO2 and comparable to the lightweight DLIO, achieving a balance between accuracy and real-time performance.

\bibliographystyle{IEEEtran}
\bibliography{references.bib}

\begin{IEEEbiography}[{\includegraphics[width=1in,height=1.25in,clip,keepaspectratio]{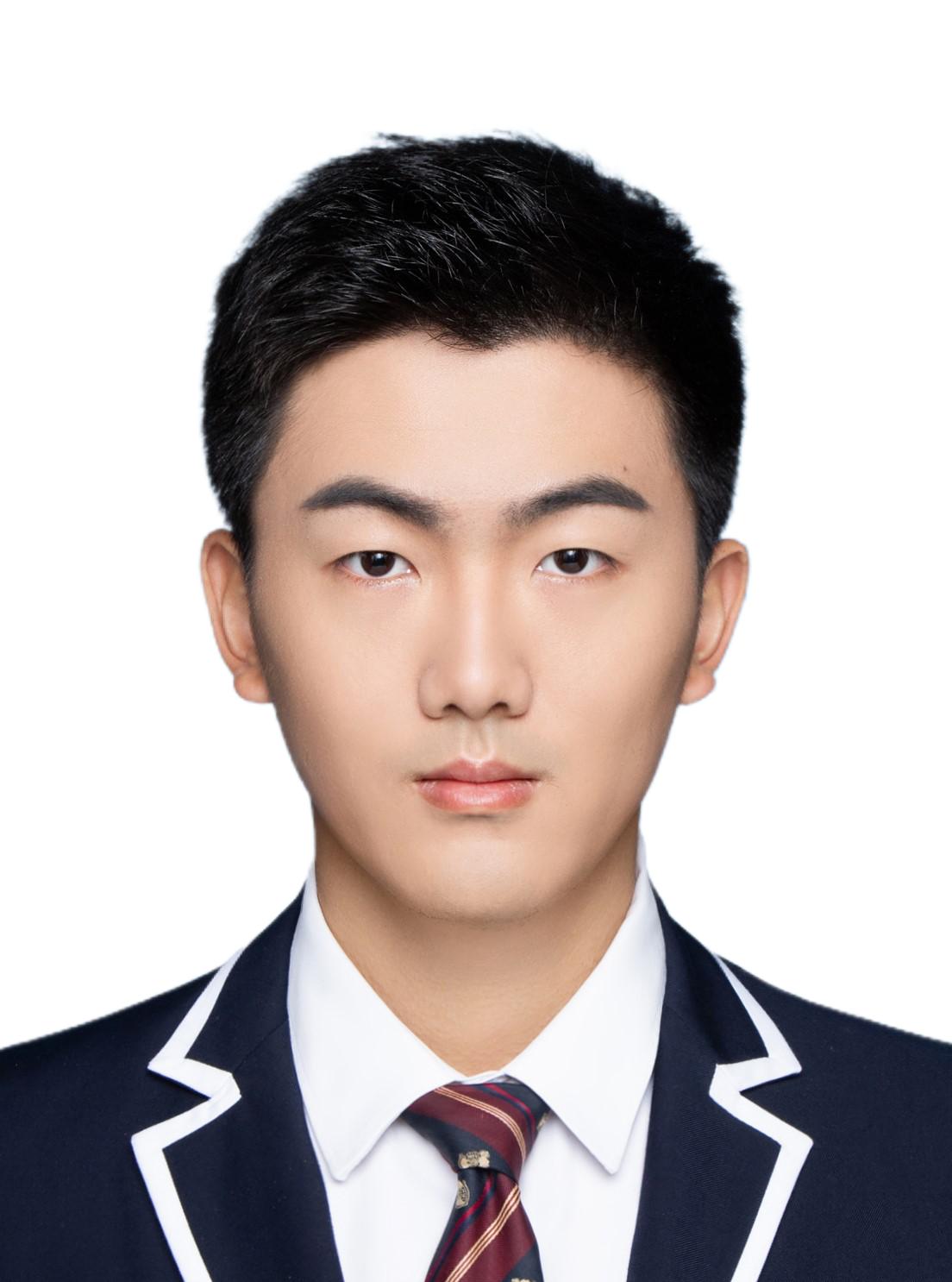}}]{Tianxiang Zhang}
(Graduate Student Member, IEEE) is pursuing the M.S. degree in Computer Application Technology at the State Key Laboratory of Surveying and Mapping Remote Sensing Information Engineering at Wuhan University, and received the B.Eng. degree in Electronic Information Engineering from Wuhan University. He was awarded the National First Prize of Intel Cup Undergraduate Electronic Design Contest - Embedded System Design Invitational Contest and designated as Outstanding Winner and AMS Award in the Interdisciplinary Contest in Modeling in 2022. His main research direction is Multi-sensor Simultaneous Localization and Mapping.
\end{IEEEbiography}
\begin{IEEEbiography}[{\includegraphics[width=1in,height=1.25in,clip,keepaspectratio]{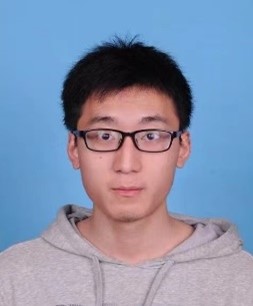}}]{Xuanxuan Zhang}
(Graduate Student Member, IEEE) obtained his bachelor's degree in Surveying and Mapping Engineering at the China University of Geosciences (Beijing) and  master's degree in Geodesy and Surveying Engineering at the Chinese Academy of Surveying and Mapping. He is currently working toward a doctoral degree in Geodesy and Surveying Engineering with the State Key Laboratory of Information Engineering in Surveying, Mapping, and Remote Sensing at Wuhan University, Wuhan, China. The main research direction is multi-sensor integrated navigation and GNSS high-precision data processing.
\end{IEEEbiography}
\begin{IEEEbiography}[{\includegraphics[width=1in,height=1.25in,clip,keepaspectratio]{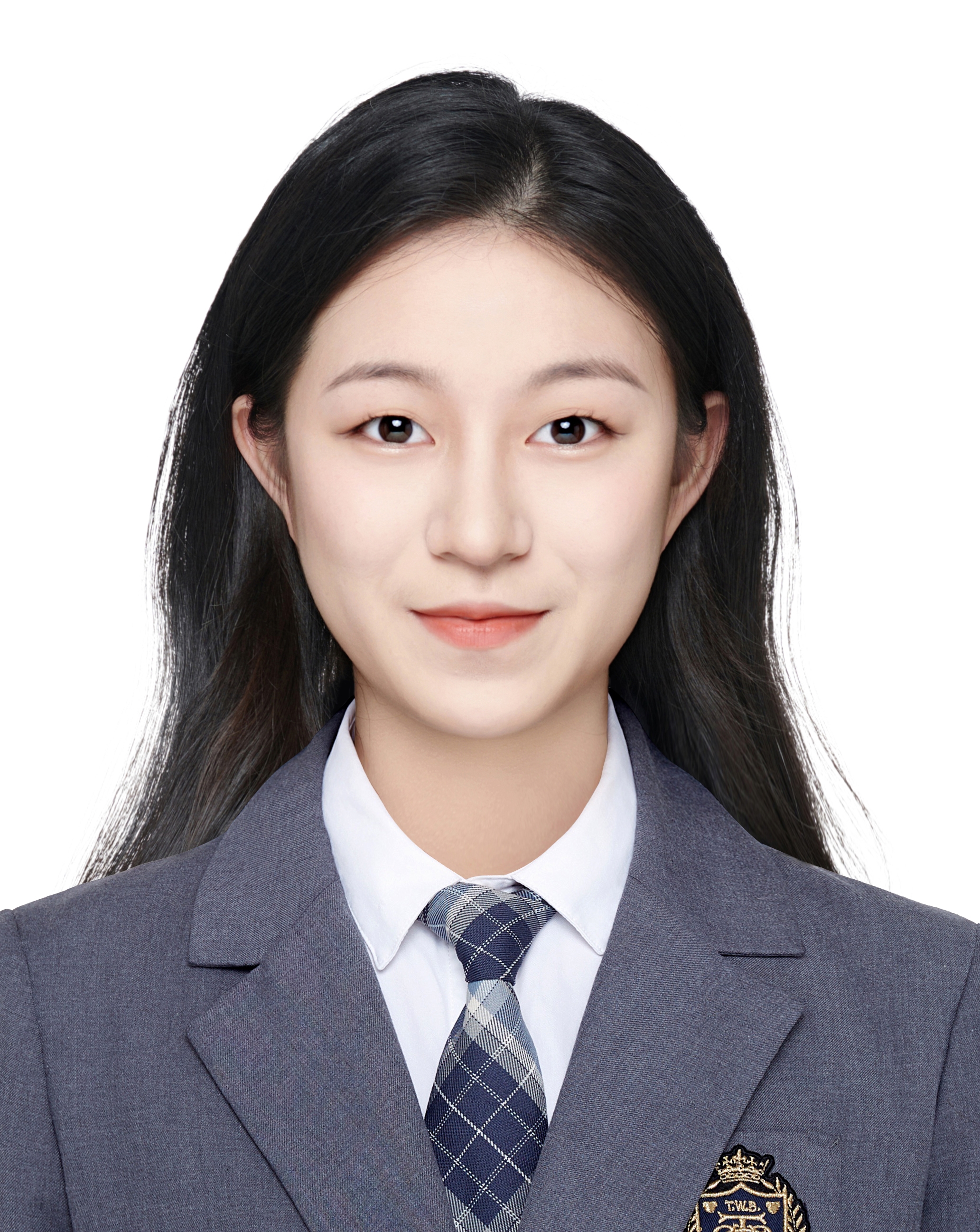}}]{Wenlei Fan}
is pursuing a Master's degree in Cartography and Geographic Information System at the State Key Laboratory of Surveying and Mapping Remote Sensing Information Engineering at Wuhan University. She received the B.S. degree in remote sensing science and technology from Wuhan University, Wuhan, China. Her research interests include point cloud processing and simplification. She was honored as a Meritorious Winner in the Interdisciplinary Contest in Modeling in 2022 and 2023. Her research interests include point cloud processing and simplification.
\end{IEEEbiography}
\begin{IEEEbiography}[{\includegraphics[width=1in,height=1.25in,clip,keepaspectratio]{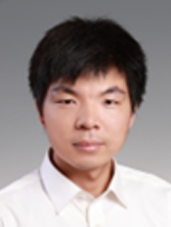}}]{Xin Xia}
(Senior Member, IEEE) received the B.E. degree in vehicle engineering from the School of Mechanical and Automotive Studies, South China University of Technology, Guangzhou, China, in 2014, and the Ph.D. degree in vehicle engineering from the School of Automotive Studies, Tongji University, Shanghai, China, in 2019. He was a Postdoctoral Fellow associated with Dr. Amir Khajepour at the Department of Mechanical and Mechatronics Engineering, University of Waterloo, Waterloo, ON, Canada, from Jan.2020 to March.2021. He is currently a Research Scientist with the Department of Civil and Environmental Engineering, University of California, Los Angeles, CA, USA. His research interest includes state estimation, cooperative localization, cooperative perception, and dynamics control of the autonomous vehicle.
\end{IEEEbiography}
\begin{IEEEbiography}[{\includegraphics[width=1in,height=1.25in,clip,keepaspectratio]{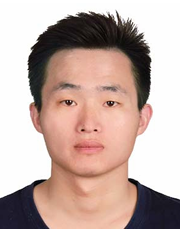}}]{Huai Yu}
(Member, IEEE) received the BS and PhD degrees in communication and information systems from Wuhan University, China, in 2015 and 2020, respectively. From 2018 to 2020 and 2020 to 2021, he has been a visiting scholar and postdoctoral fellow, respectively, with the Robotics Institute, Carnegie Mellon University, Pittsburgh, PA, USA. He is currently working as a research associate professor with the School of Electronic Information, Wuhan University. His research involves multi-modal visual feature detection and matching, structure from motion, and SLAM.
\end{IEEEbiography}
\begin{IEEEbiography}[{\includegraphics[width=1in,height=1.25in,clip,keepaspectratio]{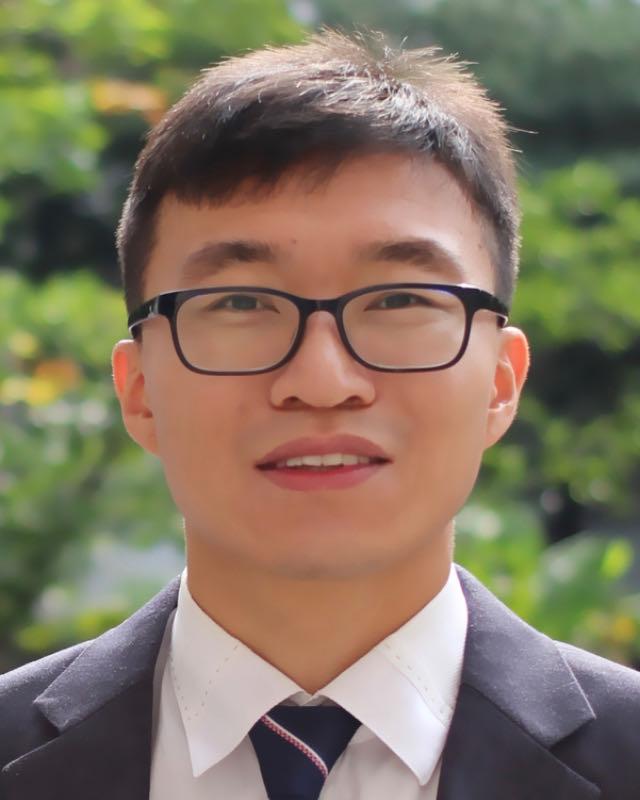}}]{Lin Wang}
(Member, IEEE) received the MS degree in CAD/CAM and VR/AR for engineering design and manufacturing, and PhD (with highest honors) degree in AI and robotics from the Korea Advanced Institute of Science and Technology (KAIST), Korea. He is currently an assistant professor with the School of Electrical and Electronic Engineering (EEE), Nanyang Technological University, Singapore. Before that, he was a faculty member with the Artificial Intelligence Thrust, HKUST-GZ, HKUST FYTRI, and Department of CSE, HKUST. He was a visiting researcher with the Department of EEE, Imperial College London (ICL), U.K. (2020–2021). He serves as a technical committee member of the IEEE Robotics and Automation Society (RAS) on robot vision. His research interests include brain-inspired machine sensing and perception, multi-modal AI, intelligent systems, etc.
\end{IEEEbiography}
\begin{IEEEbiography}[{\includegraphics[width=1in,height=1.25in,clip,keepaspectratio]{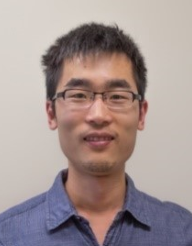}}]{You Li}
(Senior Member, IEEE)  is a Professor at the State Key Laboratory of Information Engineering in Surveying, Mapping and Remote Sensing (LIESMARS), Wuhan University, China. He is also an Adjunct Professor at the Hubei Luojia Laboratory. He received Ph.D. degrees in Geodesy and Surveying Engineering from Wuhan University and in Positioning, Navigation, and Wireless Location in Geomatics Engineering from the University of Calgary in 2015 and 2016, respectively, and a BEng degree in Surveying and Mapping Engineering from China University of Geoscience (Beijing) in 2009. His research focuses on positioning and motion-tracking techniques and applications. He has co-published over 90 academic papers and has over 20 patents pending. He serves as an Associate Editor for the IEEE Sensors Journal, a committee member at the International Association of Geodesy (IAG) unmanned navigation system session, China Society of Surveying and Mapping location services session, and the secretary at the International Society for Photogrammetry and Remote Sensing (ISPRS) mobile mapping session, and co-chairs of multiple international workshops or conferences.
\end{IEEEbiography}

\vfill

\end{document}